\documentclass[letterpaper, 10 pt, conference]{ieeeconf}  

\IEEEoverridecommandlockouts                              

\usepackage{graphicx} 
\usepackage{float} 
\usepackage{amsmath} 
\usepackage{amssymb}  
\usepackage{algorithm} 
\usepackage{algpseudocodex}
\usepackage[font={small}]{caption}
\usepackage{subcaption}
\usepackage{booktabs}
\usepackage{makecell}

\title{\LARGE \bf
Fine-Tuning Hard-to-Simulate Objectives for Quadruped Locomotion: A Case Study on Total Power Saving 
}

\author{Ruiqian Nai$^{123}$, Jiacheng You$^{123}$, Liu Cao$^{4}$, Hanchen Cui$^{3}$, Shiyuan Zhang$^{1}$, Huazhe Xu$^{123}$, and Yang Gao$^{123}$
\thanks{$^{1}$ Institute for Interdisciplinary Information Sciences, Tsinghua University. $^{2}$ Shanghai AI Lab. $^{3}$ Shanghai Qi Zhi Institute. $^{4}$ Department of Electronic Engineering, Tsinghua University.
        {\tt\small nrq22@mails.tsinghua.edu.cn}}%
}

\begin{document}

\newcommand{\rmain}{\ensuremath{{r}^\text{main}}}
\newcommand{\rhard}{\ensuremath{{r}^\text{hard}}}
\newcommand{\Rhard}{\ensuremath{{R}^\text{hard}}}
\newcommand{\yhard}{\ensuremath{{y}^\text{hard}}}
\newcommand{\pianchor}{\ensuremath{{\pi}^\text{anchor}}}
\newcommand{\pipretrained}{\ensuremath{{\pi}^\text{pre-trained}}}

\newcommand{\rms}{\ensuremath{\mathbf{s}}}
\newcommand{\rma}{\ensuremath{\mathbf{a}}}

\newcommand{\simulation}{\ensuremath{\text{sim}}}
\newcommand{\real}{\ensuremath{\text{real}}}

\maketitle
\thispagestyle{empty}
\pagestyle{empty}

\begin{abstract}
Legged locomotion is not just about mobility; it also encompasses crucial objectives such as energy efficiency, safety, and user experience, which are vital for real-world applications. However, key factors such as battery power consumption and stepping noise are often inaccurately modeled or missing in common simulators, leaving these aspects poorly optimized or unaddressed by current sim-to-real methods. Hand-designed proxies, such as mechanical power and foot contact forces, have been used to address these challenges but are often problem-specific and inaccurate. 
                        
In this paper, we propose a data-driven framework for fine-tuning locomotion policies, targeting these hard-to-simulate objectives. Our framework leverages real-world data to model these objectives and incorporates the learned model into simulation for policy improvement. We demonstrate the effectiveness of our framework on power saving for quadruped locomotion, achieving a significant 24-28\% net reduction in total power consumption from the battery pack at various speeds. In essence, our approach offers a versatile solution for optimizing hard-to-simulate objectives in quadruped locomotion, providing an easy-to-adapt paradigm for continual improving with real-world knowledge. Project page \tt{https://hard-to-sim.github.io/}.
\end{abstract}

\section{Introduction}
Legged locomotion grants robots the ability to traverse complex terrains, offering universal mobility \cite{ha2024learning}. Beyond the goal of command following, broader considerations such as energy efficiency, payload capacity, user-friendly interaction, and safety are critical for extending their utility in industrial and everyday environments \cite{biswal2021development, DUFFY2003177, breazeal2004designing, hagele2016industrial}. For quadruped robots in particular, constraints such as limited battery life per charge significantly impact the effectiveness of robots in tasks like patrolling or rescue operations \cite{unitreego1, unitreego2, katz2019mini}. Moreover, issues like disruptive stepping noise can negatively affect user experience \cite{Trovato2018Sound}, and motor overheating poses risks to the robot's operational lifespan \cite{trujillo2009thermally}.

The current successes of learning-based approaches heavily depend on the sim-to-real transfer paradigm \cite{tan2018sim,peng2018sim}, where policies trained in simulations are directly applied to real-world robots \cite{hwangbo2019learning,xiong2024adaptive,agarwal2023legged,margolis2023walk,zhuang2023robot,margolis2024rapid,peng2020learning,escontrela2022adversarial}. These methods leverage physics simulators like Isaac Gym \cite{isaacgym}, MuJoCo \cite{todorov2012mujoco}, and Bullet \cite{coumans2019bullet}, which mainly focus on dynamics and kinematics. However, critical factors like power consumption, stepping noise and safety features are not available or inaccurately modeled. These factors are hard to simulate due to the complexity of the underlying mechanisms. 
For instance, the intricate dynamics of Permanent Magnet Synchronous Motors (PMSMs) pose significant challenges in predicting total power consumption \cite{kundur2007power,mellor2009computationally}. Furthermore, the implementation of control strategies, such as Field-Oriented Control (FOC), adds another layer of complexity to accurately forecasting motor power requirements \cite{unitreego1motor,wang2018advanced}.

Traditional approaches have utilized hand-designed proxies as reward functions for training policies, applying metrics like mechanical power or foot contact forces to approximate energy consumption or noise \cite{mahankali2024maximizing, yang2022fast, fu2021minimizing, rudin2022learning, margolis2023walk}. These methods demand expert knowledge and intensive tuning, and their efficacy is limited by the accuracy of the proxies used. In contrast, learning from real-world experience offers a more precise and efficient alternative for optimizing these challenging objectives.

In this paper, we introduce a data-driven fine-tuning approach to optimize hard-to-simulate objectives in locomotion policies. Our method begins with the collection of real-world data using a pre-trained policy. We then develop a measurement model to predict hard-to-simulate factors from this data, which is integrated into the simulation as a reward function. Our approach performs iterative policy improvement through cycles of data collection and policy updating.

We present experimental results demonstrating a significant 24-28\% reduction in total power consumption from the battery pack for quadruped locomotion. The task objective—minimizing power consumption while maintaining locomotion performance—highlights the operational time constraints faced by current low-cost quadruped robots \cite{unitreego1, unitreego2, minicheetahbattery}. Despite the complexities of modeling total power consumption \cite{seok2014design,wang2018advanced}, our method effectively manages these challenges, illustrating its potential to address demanding objectives in quadruped locomotion.

We believe that our proposed framework could be applied to a wide range of hard-to-simulate objectives. It utilizes a data-driven measurement model, which is designed to be objective-agnostic and has the potential to automatically adapt to various challenging objectives. Technically, our approach requires only minimal modifications to the existing sim-to-real pipeline, offering a plug-and-play method for integrating empirical data from the physical world to enhance locomotion performance.

\begin{figure*}[t]
\centering
\includegraphics[width=0.99\linewidth]{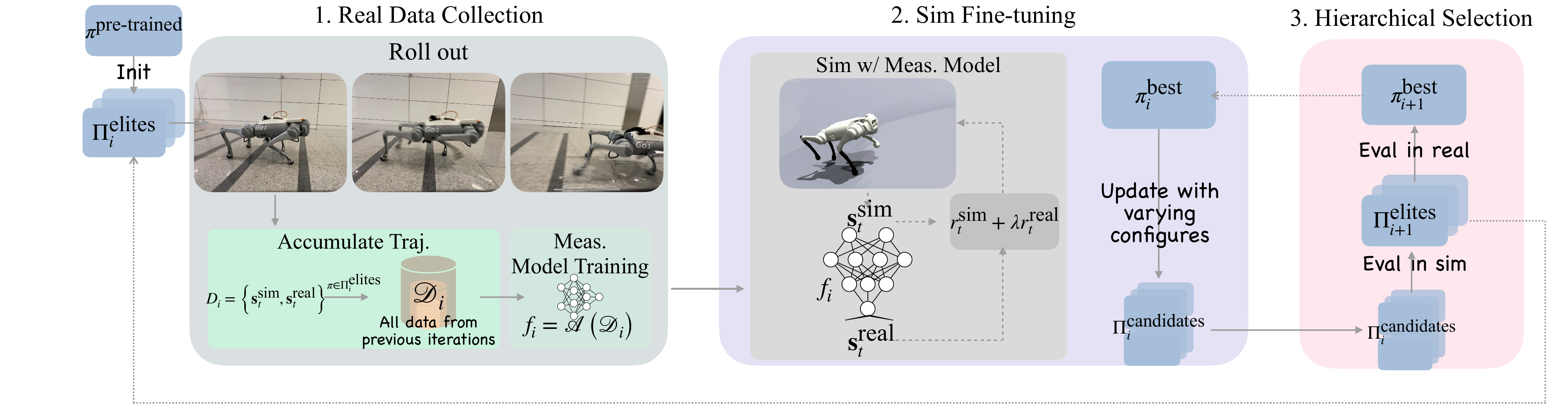}
\caption{Fine-tuning procedure at the $i$-th iteration. Real-world data is collected using the pre-trained policy or the policy batch from the last iteration. A measurement model is trained to estimate hard-to-simulate factors based on the data, which is later integrated into simulation. Policy updates are performed in simulation, generating a batch of policies with varying configurations. Hierarchical policy selection is then conducted based on performance in simulation and real-world evaluations, determining policies retained for the next iteration.}
\label{fig:method}
\end{figure*}

\begin{algorithm}[t]
\caption{Fine-tuning hard-to-simulate objectives}
\begin{algorithmic}[1]
        \footnotesize
        \Require Pre-trained policy \pipretrained, an empty real-world dataset $\mathcal{D}_0$, measurement model training algorithm $\mathcal{A}: \mathcal{D} \mapsto f$, parameter search space for $w_\cdot$ and $c$: $\mathcal{W}$ and $\mathcal{C}$, and the size of the elite policy batch: $K$.      
        \State $\Pi_0^{\text{elites}} \gets \left\{ \pipretrained \right\}$.
        \For {$i = 0,  1, 2, \ldots$}
        \State{\textcolor{gray}{// \textsc{Real-world data collection:}}}
        \State{$D_i \gets \varnothing$}
        \For {$\pi \in \Pi_i^\text{elites}$}
        \State{Roll out $\pi$ in the real world to collect data, $\left\{\rms^\simulation_t, \rms^\real_t\right\}^{\pi}$}.
        \State{${D}_i \gets D_i \cup \left\{\rms^\simulation_t, \rms^\real_t\right\}^{\pi}$}.
        \EndFor
        \State{Accumulate all real data: $\mathcal{D}_i \gets \bigcup_{j=0}^{i} {D}_j$}.
        \State{Train a measurement model: $f_i \gets \mathcal{A}(\mathcal{D}_i)$}.
        \State{\textcolor{gray}{// \textsc{Simulation fine-tuning:}}}
        \State{$\Pi^\text{candidates}_i \gets \varnothing $}
        \For {$\pianchor, w_\cdot, c \in \left\{\pi_i^{\text{best}}, \pipretrained \right\} \times \mathcal{W} \times \mathcal{C} $} \label{alg:main:search}
        \State{$\pi^\text{fine-tuned} \gets$ fine-tune with the objective in Eq. \ref{eq:opt-sim}}.
        \State{$\Pi_i^\text{candidates} \gets \Pi_i^\text{candidates} \cup \left\{\pi^\text{fine-tuned}\right\}$}.
        \EndFor
        \State{\textcolor{gray}{// \textsc{Hierarchical policy selection:}}}
        \State{$\Pi^\text{elites}_{i+1} \gets$ top-$K$ of $\Pi^\text{candidates}$ measured by simulation performances}. 
        \State{$\pi^{\text{best}}_{i+1} \gets   $ best of $\Pi^\text{elites}_{i+1}$ evaluated by real-world performances}.
        \EndFor
\end{algorithmic}
\label{alg:main}
\end{algorithm}
\section{Related Work}

\paragraph{Direct learning from real-world experience}
Training policies directly in the real world can circumvent the complexities associated with modeling hard-to-simulate factors. However, previous methods \cite{ha2020learning, haarnoja2018learning, smith2022walk, smith2023grow} typically achieve results only in low-performance regions characterized by slow walking speeds. On the other hand, methods that fine-tune simulation-trained policies in real-world \cite{smith2022legged, lei2023uni, shi2024efficient} primarily addresses the shift in dynamics due to sim-to-real transferring. These studies often do not extend to optimizing objectives that are unseen in simulation pre-training. Furthermore, in contrast to these methods, which are generally tailored for specific learning algorithms \cite{hiraoka2021dropout, shao2020controlvae}, our approach offers a versatile integration across various policy optimization frameworks.

\paragraph{Augmenting simulation with data-driven models} 
Recent advances have embraced hybrid simulations, combining analytic physics with data-driven models to capture complex dynamics \cite{heiden2021neuralsim,qiao2021efficient,heiden2022inferring,sanchez2020learning,ajay2018augmenting}. Such hybrid simulations find applications in diverse robotic tasks, including legged locomotion \cite{hwangbo2019learning, margolis2023walk}, drone racing \cite{kaufmann2023champion},  and modeling human behavior in sports \cite{abeyruwan2023sim2real}. These studies, however, primarily focus on enhancing the fidelity of simulation \textit{dynamics}. In contrast, our setting requires data-driven models to be explicit \textit{objectives} for policy optimization, which exacerbates issues related to out-of-distribution and model exploitation.

\paragraph{Energy efficiency in quadruped robots} 
Total energy efficiency optimizations are typically reserved for robot hardware design \cite{seok2014design, katz2019mini, krimsky2024elastic}. On the controller side, previous works have employed various strategies to estimate and optimize energy consumption in legged locomotion. Techniques include using mechanical power and Joule heating as reward functions \cite{hwangbo2019learning, rudin2022learning, margolis2024rapid, margolis2023walk, agarwal2023legged, zhuang2023robot, yang2022fast, fu2021minimizing} or as constraints \cite{mahankali2024maximizing, kim2024not, chane2024cat}. Rather than relying on hand-designed proxies, our approach aims to minimize total power consumption through data-driven fine-tuning techniques.

\section{Fine-tuning Hard-to-Simulate Objectives}
\subsection{Problem Formulation}
We model locomotion as a Markov Decision Process (MDP) with state space $\mathcal{S} \subset \mathbb{R}^n$, action space $\mathcal{A} \subset \mathbb{R}^m$, transition function $p(\cdot \mid \rms, \rma) : \mathcal{S} \times \mathcal{A} \times \mathcal{S} \rightarrow [0,1]$, and reward function $r: \mathcal{S} \rightarrow \mathbb{R}$. The objective is to find a policy $\pi : \mathcal{S} \rightarrow \mathcal{A}$ that maximizes the expected sum of discounted rewards $\mathbb{E}\left[\sum_{t=0}^T \gamma^t r(\rms_t)\right]$, where $\gamma$ is the discount factor and $T$ is the time horizon. To address hard-to-simulate factors, we divide the state space into $\mathcal{S} = \mathcal{S}^\simulation \times \mathcal{S}^\real$, where $\mathcal{S}^\simulation$ represents states available in simulation and $\mathcal{S}^\real$ represents states observed only in the real world, i.e., the hard-to-simulate factors. 

We formulate the locomotion task as a multi-objective optimization problem \cite{sener2018multi}, highlighting objectives beyond mobility. For example, besides following velocity commands, the robot should minimize power consumption, ensure safety, and interact friendly with humans. To this end, we implement linear scalarization \cite{gunantara2018review} to combine multiple objectives into a single reward function, following common practice \cite{hwangbo2019learning, rudin2022learning, margolis2023walk}.
Specifically, we factorize the reward as:
$ r(\rms_t) = \sum_i w_i r_i^\simulation(\rms^\simulation_t) + \sum_j w_j r_j^\real(\rms^\real_t)$,
where $w_\cdot$ are the weighting factors for each objective. To summarize, the optimization problem of fine-tuning hard-to-simulate objectives is:
\begin{equation}
\begin{aligned}
\max_{\pi} & \mathop{\mathbb{E}}_{
        \substack{
                \rma_t \sim \pi(\cdot \mid \rms_t) \\
                \rms_{t+1} \sim p^\real(\cdot \mid \rms_t, \rma_t)
        } 
} \left[ \sum_{t=0}^T \gamma^t r(\rms_t)\right], \\
\text{where}  \quad & r(\rms_t) = \sum_i w_i r_i^\simulation(\rms^\simulation_t) + \sum_j w_j r_j^\real(\rms^\real_t).
\end{aligned}
\label{eq:opt}
\end{equation}
Note that our ultimate goal is to optimize real-world performance. Therefore, the expectation here is taken over the real-world transition function $p^\real$. And the rewards related to hard-to-simulate factors,   $r^\real_\cdot$, are calculated based on real-world measurements $\rms^\real_t$.

\subsection{Algorithm Design and Motivation}
The fine-tuning process is depicted in Figure~\ref{fig:method} and detailed in Algorithm~\ref{alg:main}. We employ a data-driven approach to model hard-to-simulate factors, training a measurement model to estimate these factors from real-world data. To tackle the issues of out-of-distribution data and potential model exploitation during fine-tuning, we impose constraints on the size of policy updates, as discussed in Section~\ref{sec:measurement}. Considering the incremental improvements from each update, we iteratively conduct policy learning and measurement model training. In each cycle, we sweep through multiple training configurations to adapt to the evolving characteristics of the measurement model (see Section~\ref{sec:iterative}). At the conclusion of each iteration, we systematically select the most effective policies through a hierarchical process, initially in simulation followed by real-world evaluations, as described in Section~\ref{sec:hierarchical}.

Our fine-tuning methodology consists of three steps: gathering real-world data, updating policies through simulation, and hierarchically selecting the most effective policies. This iterative process continues until the desired performance metrics are achieved.

\subsection{Data-driven Measurement Model}\label{sec:measurement}
To address the hard-to-simulate factors, we develop a measurement model that is trained end-to-end using real data. This model predicts these factors from observations available in simulation, denoted as $f: \rms^\simulation \mapsto \widehat{\rms^\real}$. 

The training of the measurement model is straightforward: we deploy a trained policy and collect pairs of observations and hard-to-simulate factors, $\mathcal{D}=\left\{\rms^\simulation_t, \rms^\real_t\right\}$. We then train the model $f$ with an algorithm $\mathcal{A}: \mathcal{D} \mapsto f$. E.g., to minimize the prediction error, $\mathbb{E}_{\mathcal{D}}\left[\left\| f(\rms^\simulation_t) - \rms^\real_t \right\|^2\right]$, on the real dataset. Subsequently, we integrate the measurement model into the simulation to generate rewards, $r^\real\left(f(\rms^\simulation_t)\right)$, for optimizing hard-to-simulate objectives.

However, the distribution of the training data for the measurement model is heavily dependent on the data-collecting policy. As the policy deviates from the data-collecting policy during fine-tuning, prediction accuracy may decrease due to out-of-distribution issues. Furthermore, since the model serves as an explicit objective, the optimization algorithm may exploit it. To address this, we constrain the policy update step size using a KL divergence penalty \cite{schmitt2018kickstarting}. This penalty encourages the policy to stay close to the data-collecting policy, ensuring the measurement model's reliability.

Therefore, the objective for policy optimization in simulation is:
\begin{equation}
\begin{aligned}
    \max_{\pi} \; & \mathop{\mathbb{E}}_{
            \substack{
                \rma_t \sim \pi(\cdot \mid \rms_t) \\
                \rms_{t+1} \sim \textcolor{gray}{p^\simulation(\cdot \mid \rms_t, \rma_t)}
            } 
    } \left[ \sum_{t=0}^T \gamma^t r(\rms_t) \right], \\
    \text{s.t.} \; & \mathop{\mathbb{E}}_{\rms \sim \pi}\left[\text{KL}\left(\pi(\cdot \mid \rms) \parallel \pianchor(\cdot \mid \rms)\right) \right] \leq c, \\
    \text{where} \; & r(\rms_t) = \sum_i w_i r_i^\simulation(\rms^\simulation_t) + \sum_j w_j \textcolor{cyan}{r_j^\real\left(f(\rms^\simulation_t)\right)}.
\end{aligned}
\label{eq:opt-sim}
\end{equation}
Here, \pianchor represents the policy initialization for fine-tuning, which is detailed in Section \ref{sec:iterative}. The KL divergence penalty is controlled by the parameter $c$. The hard-to-simulate factors are estimated using the measurement model $f$.

\subsection{Iterative Policy Fine-tuning} \label{sec:iterative}
Policy improvement is often limited by constraints on the update step size. To address this limitation, we employ an iterative approach that involves collecting real data and updating the policy multiple times. Additionally, we aggregate all data collected in previous iterations to train the measurement model, enhancing data diversity and mitigating the out-of-distribution issue \cite{abeyruwan2023sim2real}. This iterative process benefits both the modeling of hard-to-simulate factors and policy optimization, leading to improved overall performance.

Our empirical findings indicate that iterating with fixed hyper-parameters is suboptimal. This is because the appropriate weighting factors $w_\cdot$ and the KL divergence penalty $c$ depend on the characteristics of the measurement model $f$, and their optimal values may shift as the model evolves. Consequently, we adapt our policy learning approach in each iteration by varying configurations, sweeping through different combinations of $w_\cdot$ and $c$. Additionally, for the policy initialization, \pianchor, we vary it between the pre-trained policy \pipretrained{} and the best policy from the previous iteration $\pi^{\text{best}}$. Resetting the policy to \pipretrained{}—but updated with the improved measurement model $f$—provides extra flexibility and adaptation capabilities \cite{nikishin2022primacy,schwarzer2023bigger,smith2023grow}.

\subsection{Hierarchical Policy Selection}\label{sec:hierarchical}

Policy fine-tuning in simulation using parameter sweeping generates multiple policy candidates. Deciding which policies to retain for subsequent iterations poses a challenge due to the sim-to-real gap, which leads to discrepancies between real-world and simulation performances. Formally, differences between $p^\real$ and $p^\simulation$ lead to variations in the objectives, as denoted by Eq.~\ref{eq:opt} and Eq.~\ref{eq:opt-sim}. Therefore, we propose a hierarchical policy selection method: first, we select the top-$K$ policies in simulation, and then we further select the best-performing policy in the real world.

Initially, we select a set of elite policies, ${\Pi}^\text{elites}$, based on their performance in simulation. Subsequently, we deploy ${\Pi}^\text{elites}$ for real-world evaluations. The top-performing policy in the real world, $\pi^{\text{best}}$, is then designated as the anchor policy for the next iteration. Concurrently, the real-world data collected from ${\Pi}^\text{elites}$ is incorporated into $\mathcal{D}$, further enhancing the training of the measurement model.

\section{Experimental Design}

To evaluate the efficacy of our framework, we concentrate on optimizing the challenging objective of total power savings. This metric is crucial in real-world applications as it determines the operating duration per charge for quadruped robots \cite{unitreego1, unitreego2, katz2019mini}. The accurate simulation of total power consumption poses significant challenges due to the intricate energy interactions among different subsystems \cite{seok2014design} and the complex, time-variant dynamic of Permanent Magnet Synchronous Motors (PMSMs) \cite{unitreego1motor,kundur2007power,mellor2009computationally}.

Conventionally, optimization for power saving focuses on hand-designed proxies that represent the \textit{analytical} power consumption, including mechanical power and Joule heating. 
In contrast, our approach targets \textit{total} power consumption, which refers to the energy drawn directly from the battery pack.

\subsection{Implementation of Our Framework and the Baseline}
\label{sec:implementation}
To estimate total power consumption, we develop a data-driven model to predict instantaneous currents $i_t$, as the voltage remains stable within a test run. Inputs to the model are motor torques and angular velocities, and we use an LSTM network, \( f \), for predictions: \( \widehat{i_t} = f(\tau_t, \dot{q}_t) \). All features and output currents are normalized using their means and standard deviations.

The locomotion policy is initially pre-trained in simulation environments with reward functions from \cite{rudin2022learning} and terrain and command curricula from \cite{rudin2022learning} and \cite{margolis2024rapid}. For fine-tuning, the real-world objective in Eq. \ref{eq:opt-sim} focuses on power saving, defined as $r^\real\left(\widehat{i_t}\right) = -\widehat{i_t}$. We apply the reward centering technique \cite{naik2024reward} to deal with statistical shifts of measurement models. The current estimation model $f$ is integrated into the simulation for current predictions, and policy updates are implemented using the Proximal Policy Optimization (PPO) algorithm \cite{schulman2017proximal}.

We simplify the combination of reward functions as \( r = r^\simulation + \lambda r^\real \). In each iteration, we adjust the reward weightings, \( \lambda \in \{0.5, 1, 5\} \), and the KL divergence penalty, \( c \in \{0.2, 0.5, 1\} \). The top-6 policies, ranked by their unweighted mean energy reward in simulation, are selected as the elite set $\Pi^{\text{elites}}$. These \(\Pi^{\text{elites}}\) will be used for real-world evaluations. The best-performing policy (\( \pi^{\text{best}} \)) in the real world--measured by total power consumption--becomes the anchor for the subsequent iteration. Data is concurrently recorded during evaluation to train the measurement model.

Conversely, the baseline fine-tunes an identical pre-trained policy in simulation but with an analytical energy reward: $r^\real \left(\rms^\simulation_t\right) = -\sum_{i=0}^{12}\max \left(\tau_i \dot{q}_i + \frac{r}{k^2}\tau_i^2, 0\right) $, where $\tau_i$ and $\dot{q}_i$ denote the torque and angular velocity of the $i$-th motor, and $r,k$ are motor constants \cite{yang2022fast,mahankali2024maximizing}. The simulation setup and policy optimization procedures mirror those used in our framework's fine-tuning phase, differing only in the energy reward function.

The hyper-parameter search space for the baseline is expanded to \( \lambda \in \{5 \times 10^{-5}, 1 \times 10^{-4}, 5 \times 10^{-4}, 1 \times 10^{-3}, 5 \times 10^{-3}\} \) and \( c \in \{0.1, 0.2, 0.5, 1.0, 5.0, \infty\} \). We fine-tune the policy using the same number of samples accumulated across all iterations of our framework. Subsequently, the top 24 policies, evaluated by the analytical energy metric, are tested in the real world, and we report the highest power reduction observed. Note that the number of real-world evaluated policies for the baseline is equal to that of our framework (\(6 \text{ elite} \times 4 \text{ iteration}\)).

\subsection{Power Measurement and Metric}
\label{sec:power-measurement}
We assess the framework by deploying the trained policies on a Unitree Go1 robot, which is commanded to maintain a constant forward velocity \( v \). Power consumption is monitored by measuring the battery's current draw at 50 Hz using the Unitree SDK, while the robot's speed is tracked using an Intel RealSense T265 camera. Each policy's power usage is evaluated over 160 seconds by calculating and averaging the current integrals from all 1-second segments that meet the velocity criteria (within $\pm 10\%$ of $v$).

For fair comparisons, we conduct direct `head-to-head' tests between the pre-trained and fine-tuned policies. These policies are alternated every 80 seconds without delay on the same robot to control for measurement variability and environmental factors. 

The primary metric is the percentage reduction in power consumption of the fine-tuned policy compared to the pre-trained one, given by \( \Delta P = \frac{P^\text{pre-trained} - P^\text{fine-tuned}}{P^\text{pre-trained}} \times 100\% \). We report both the \textit{gross power reduction} and the \textit{net power reduction}, with the latter adjusting to isolate the power consumed by locomotion processes. Specifically, we subtract the power measured when all motors are idle.

We report power reduction results at multiple speeds: $v=0.5, 0.8, 1.1$ m/s. However, for hierarchical selection during iterations, we only consider performances at \(v = 0.8\) m/s. The performance at other speeds is evaluated using the final policy after the fine-tuning process is completed.

\begin{table}[t]
\centering        
\def\arraystretch{1.2}
\begin{tabular}{l|ccc}
\toprule
 & $v=0.5$m/s & $v=0.8$m/s & $v=1.1$m/s \\ \hline
\makecell{Analytical \\ proxy} & 11.8\% (8.3\%) & 6.2\% (4.5\%) & 5.0\% (3.9\%) \\ 
\hline
\makecell{Data-driven \\ proxy  (ours)} & \textbf{28.4\% (19.6\%)} & \textbf{27.0\% (20.3\%)} & \textbf{24.2\% (19.4\%)} \\ 
\bottomrule
\end{tabular}
\caption{Comparison of net (gross) power reduction between our framework with data-driven modeling and the baseline with the analytical proxy.}
\label{tab:final-results}
\vspace{-2em}
\end{table}

\begin{figure*}[t]
\begin{subfigure}{0.15\linewidth}
\centering
\includegraphics[width=0.99\linewidth]{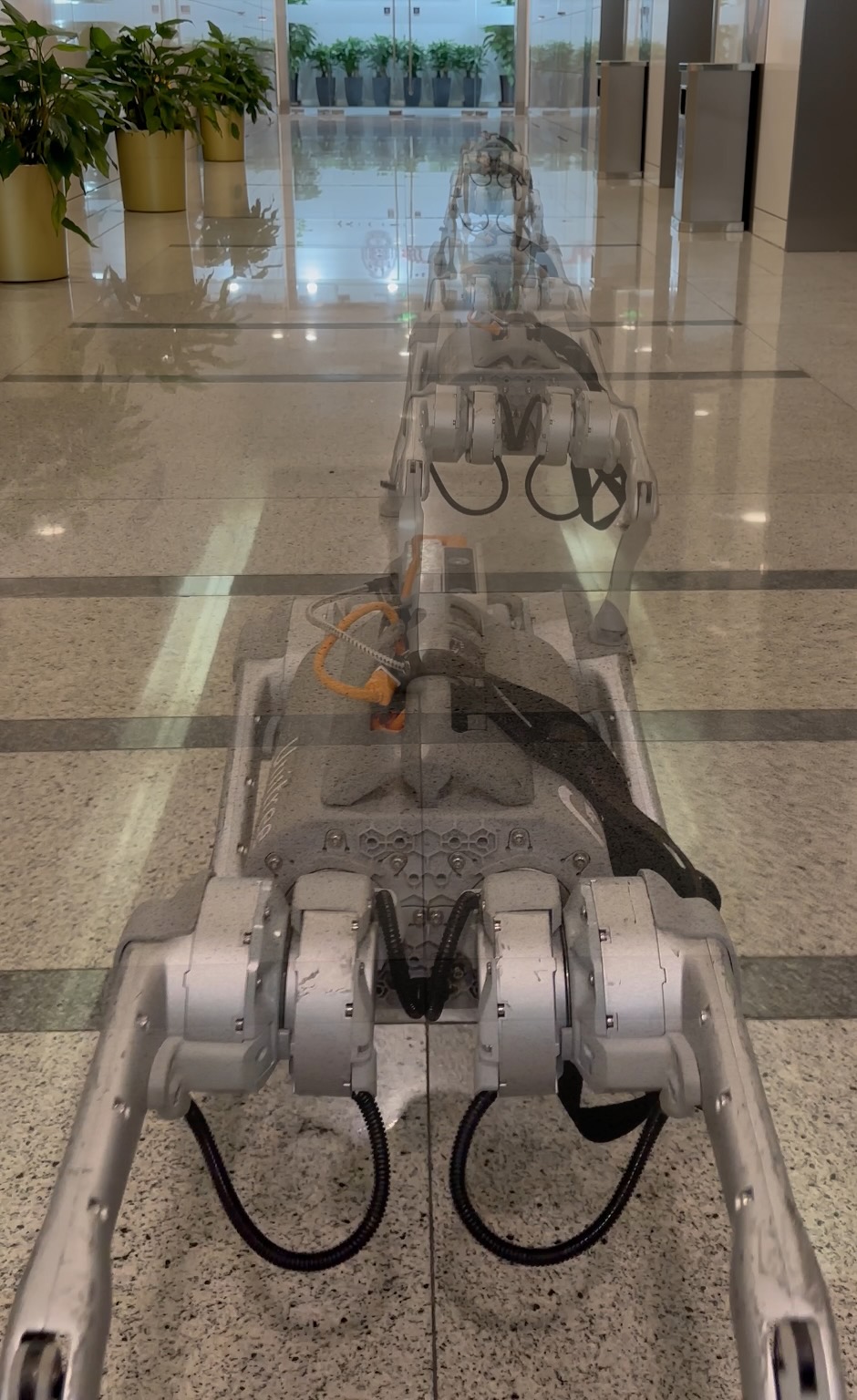}
\caption{Indoor evaluation.}
\label{fig:indoor-env}
\end{subfigure}
\begin{subfigure}{0.33\linewidth}
\centering
\includegraphics[width=0.99\linewidth]{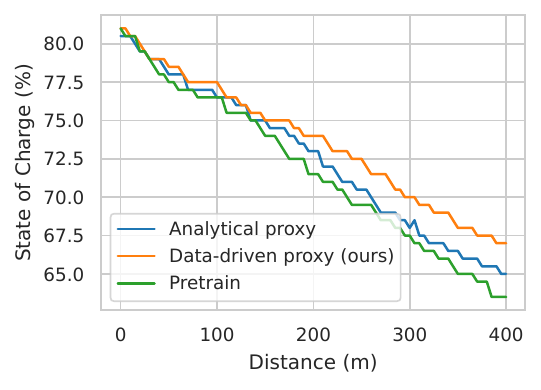}
\vspace{-1.8em}
\caption{SoC during the in-door long-distance evaluations (averaged over 3 runs). }
\label{fig:indoor-soc}
\end{subfigure}
\begin{subfigure}{0.15\linewidth}
\centering
\includegraphics[width=0.99\linewidth]{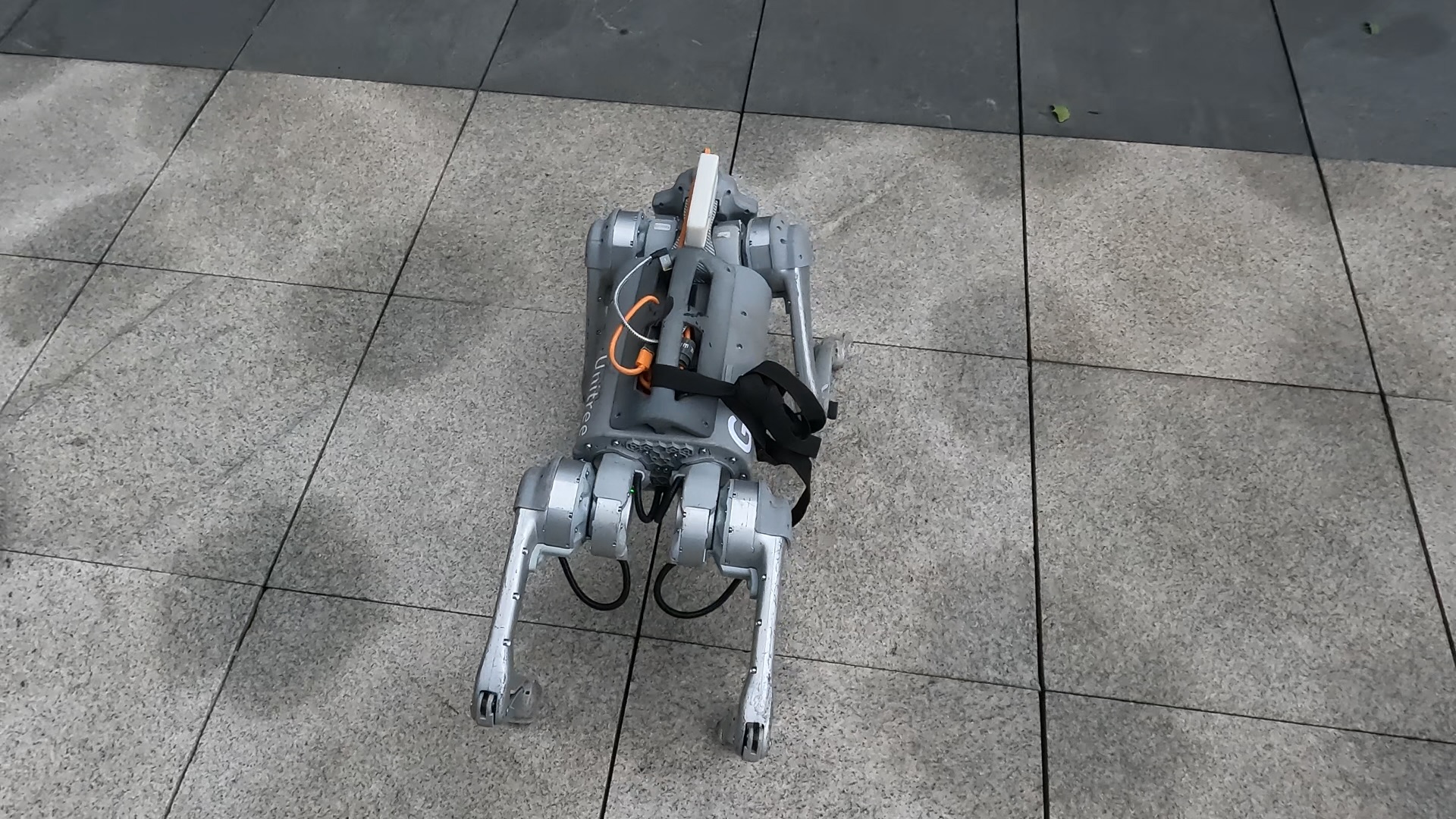}
\includegraphics[width=0.99\linewidth]{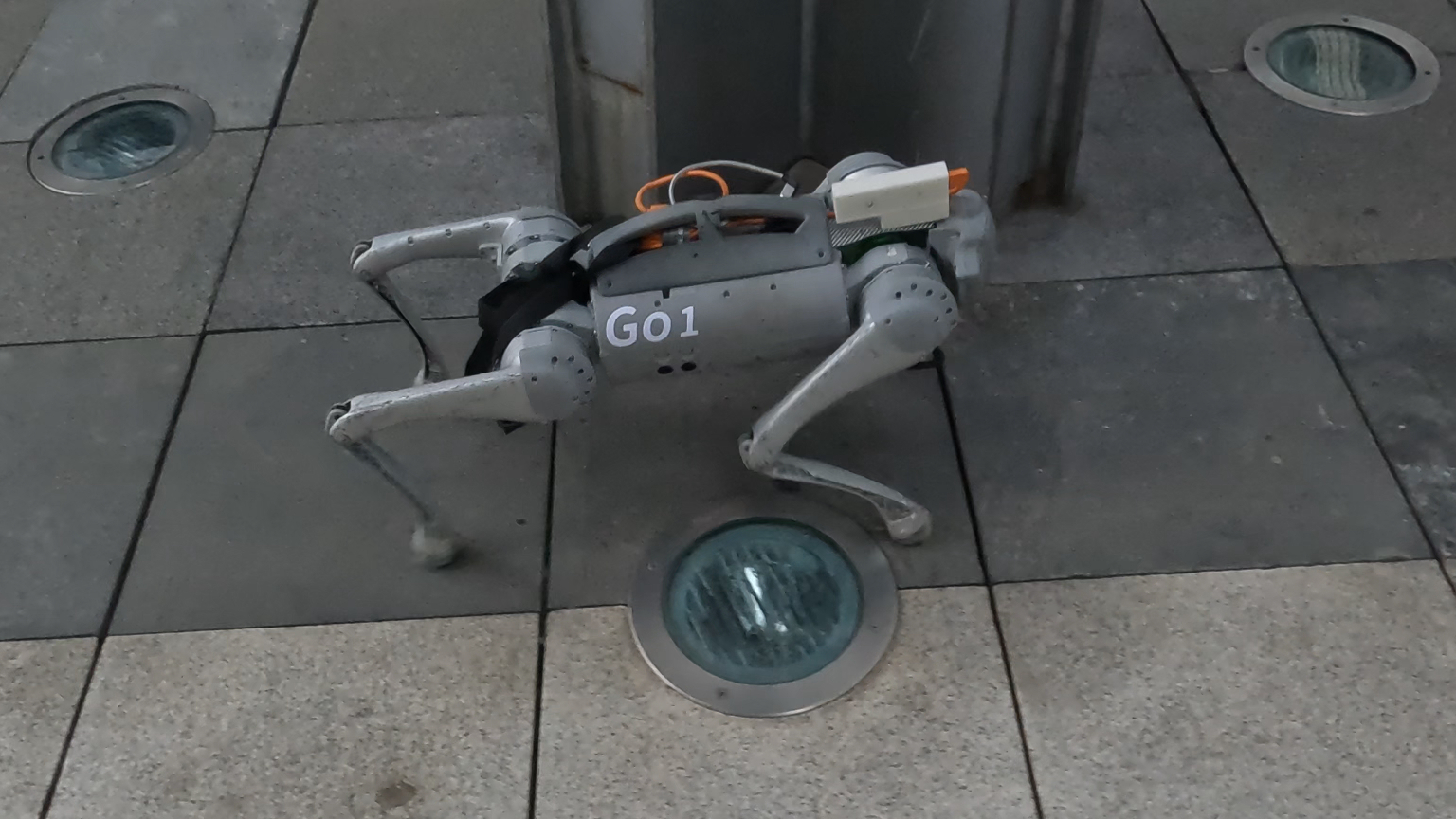}
\includegraphics[width=0.99\linewidth]{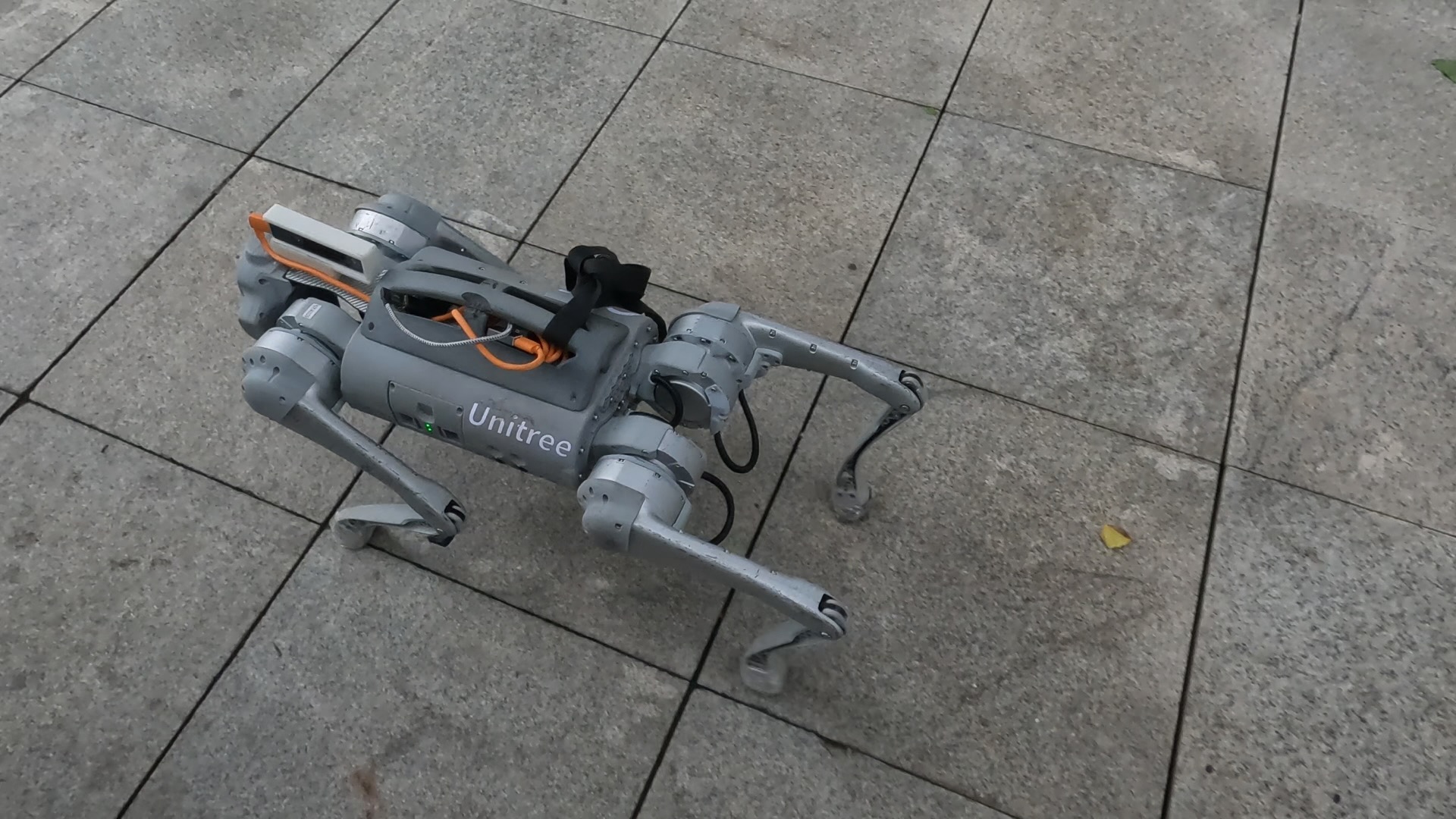}
\caption{Outdoor evaluation.}
\label{fig:outdoor-env}
\end{subfigure}
\begin{subfigure}{0.35\linewidth}
\centering
\includegraphics[width=0.99\linewidth]{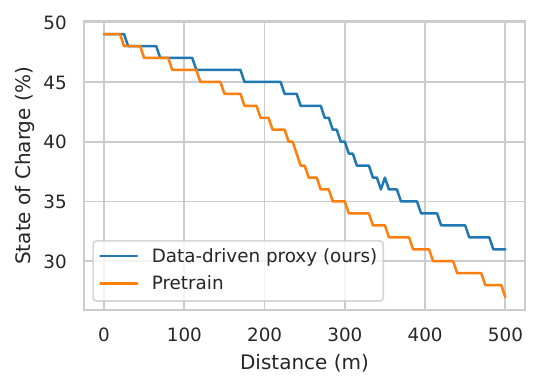}
\caption{SoC during the outdoor long-distance evaluations. }
\label{fig:outdoor-soc}
\end{subfigure}
\caption{In-the-wild evaluation environments and battery life improvements.}
\vspace{-2em}
\label{fig:in-the-wild}
\end{figure*}

\begin{figure}[t]
\centering
\includegraphics[width=0.79\linewidth]{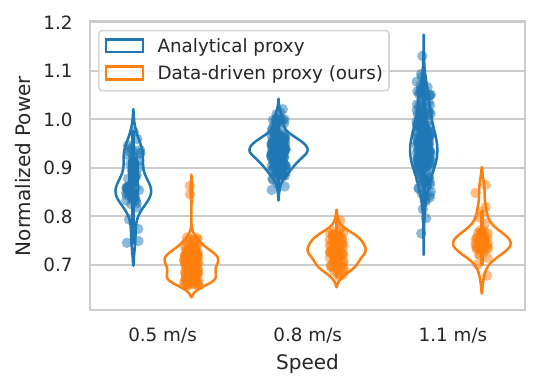}
\caption{Distribution of normalized net power ($\frac{P^\text{fine-tuned}}{\operatorname{Mean}\left(P^\text{pre-trained}\right)}$) for the policies fine-tuned with the baseline  and our framework. Each point represents a 1-second segment within the test run.}
\label{fig:power-dist}  
\vspace{-2em}
\end{figure}


\section{Main Results: Power Reduction Comparison}
We aim to demonstrate the effectiveness of our framework in optimizing hard-to-simulate objectives, focusing on total power savings. We first present the quantitative results of the metric described in Section~\ref{sec:power-measurement}, showing the net and gross power reductions achieved by our framework and the baseline. Additionally, we discuss qualitative insights into the robot's behavior changes. Finally, we conduct an in-the-wild study, evaluating battery life improvements in both indoor and outdoor long-distance tests.

\paragraph{Quantitative Results}
The final results, as summarized in Table~\ref{tab:final-results}, demonstrate that our framework achieves a significant net power reduction of 24--28\% compared to the pre-trained policy. This reduction is noteworthy, even when considering the gross power, which includes power consumption not directly related to locomotion, with our framework achieving a 20\% decrease. In contrast,  the baseline—despite thorough parameter tuning and policy selection (see Section~\ref{sec:implementation}) —still falls short, exhibiting only a 12\% net reduction at a lower speed ($v = 0.5$ m/s) and a marginal power saving (about 5\%) at higher speeds.

Figure~\ref{fig:power-dist} further illustrates these points by showing the integrated power consumption for each 1-second segment of the test run. The distribution of normalized net power, defined as $\frac{P^\text{fine-tuned}}{\operatorname{Mean}(P^\text{pre-trained})}$, highlights a consistent reduction in power consumption across all segments when using our fine-tuned policies. Conversely, the baseline policies display a more variable distribution, with some segments even surpassing the power consumption levels of the pre-trained policy.

\begin{figure}[t]
\centering
\begin{subfigure}{0.7\linewidth}
\centering
\includegraphics[width=0.99\linewidth]{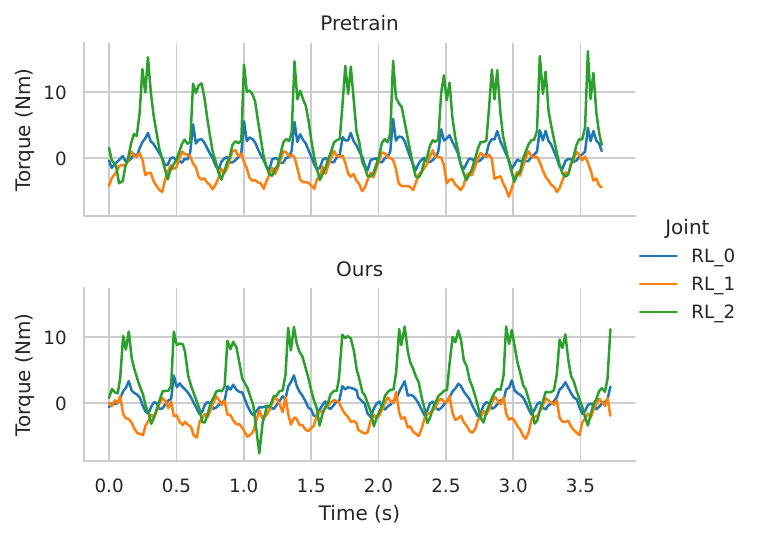}
\caption{Joint torque profiles of the rear-right leg for pre-trained and fine-tuned policies.}
\label{fig:torque}
\end{subfigure}
\hfill
\begin{subfigure}{0.27\linewidth}
\centering
\includegraphics[width=0.99\linewidth]{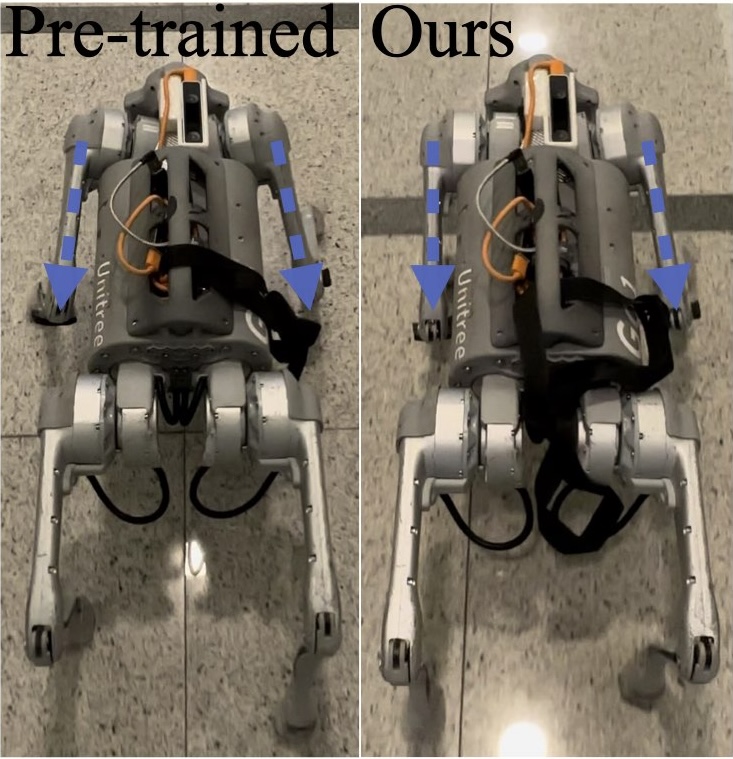}
\caption{The fine-tuned policy (right) yields a more natural front leg stance compared to the pre-trained policy (left).}
\label{fig:front-leg}
\end{subfigure}
\caption{Comparison of joint torque profiles and front leg stance between pre-trained and fine-tuned policies.}
\vspace{-2em}
\end{figure}

\paragraph{Behavioral Changes}
After fine-tuning with our framework, the robot exhibited increased compliance in its walking gait (see the accompanying video). Figure~\ref{fig:torque} illustrates the torque profiles of the rear-right leg's joints during a 0.8~m/s walk. The fine-tuned policy demonstrates smoother torque transitions and reduced amplitudes at critical joints. Notably, the front legs shifted from an outward-pointing stance to a position closer to the body (see Figure~\ref{fig:front-leg}). These adaptations indicate that the fine-tuned policy is effective in uncovering behaviors that enhance energy efficiency.


\paragraph{In-the-Wild Study}
To replicate conditions resembling realistic, uncontrolled usage scenarios, we conducted in-the-wild evaluations both indoors and outdoors, focusing on the battery's state-of-charge (SoC) relative to the distance traveled. The indoor evaluation involved a 400-meter journey around a tiled corridor (see Figure~\ref{fig:indoor-env}), while the outdoor evaluation encompassed a 500-meter route around a platform among office buildings (see Figure~\ref{fig:outdoor-env}). These long-distance evaluations required the policy to effectively manage energy over extended periods.

We initiated these tests by first fully charging the battery pack, then discharging it to 81\% for indoor tests and 49\% for outdoor tests. Starting from a fully charged battery was not feasible due to the time required to boot the programs and set up the hardware. The robot was controlled using a PID controller to maintain constant speeds. For the indoor evaluations, the target speeds were uniformly sampled from $v \in [0.5, 1.1]$ m/s, with speed adjustments every 20 seconds. For outdoor evaluations, the robot maintained a constant speed of 0.8 m/s.

Results, illustrated in Figure~\ref{fig:in-the-wild}, indicate that our fine-tuned policy maintained the highest final SoC under both indoor and outdoor conditions. The SoC of all tested methods started similarly, but as the distance traveled increased (beyond 100 meters), our fine-tuned policy exhibited a slower decline. These findings suggest that our framework not only conserves power under less controlled real-world conditions but also effectively extends battery life in more realistic usage scenarios.

\paragraph{Summary}
Our proposed framework effectively optimizes hard-to-simulate objectives, resulting in significant power savings across a range of commanded speeds. This optimization is reflected in the increased compliance observed in the robot's behaviors. Moreover, our in-the-wild evaluations further demonstrate the practical benefits of our framework, showcasing enhanced battery life in real-world scenarios.

\begin{figure}[t]
\centering
\includegraphics[width=0.7\linewidth]{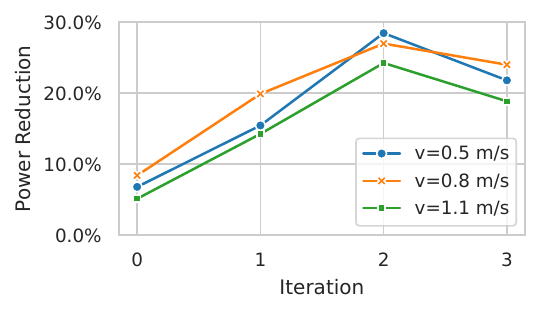}
\caption{Net power reduction over iterations. }
\vspace{-2em}
\label{fig:iterative-net}
\end{figure}

\section{Analysis}
\label{sec:analysis}
In this section, we analyze the core components of our framework. We begin with a detailed discussion on the iterative fine-tuning process, followed by a comparative analysis between our data-driven model and the analytical proxy.

\paragraph{Iterative Fine-tuning Process}
Figure~\ref{fig:iterative-net} illustrates the net power reduction achieved by our framework across successive iterations. The learning curve shows steady incremental improvements during the first three iterations, which supports the hypothesis that performance benefits from the accumulation of real-world data. This trend demonstrates robustness across all tested operational speeds. We attribute the effective adaptation of hyper-parameters to a finely balanced update step size and the reliability of the measurement model.

Convergence is achieved after four iterations, equivalent to 384,000 real-world samples (calculated as $4 \times 6 \times 16,000$). The final iteration shows a slight decrease in performance, likely due to the staleness of earlier samples. These samples, generated by older versions of the policy, become less informative over time. Consequently, the measurement model allocates capacity to memorizing these outdated samples, ultimately reaching its limit. Nonetheless, having achieved the desired level of power reduction, we conclude the iterative process at this point.

\begin{figure}[!t]
\centering
\includegraphics[width=0.89\linewidth]{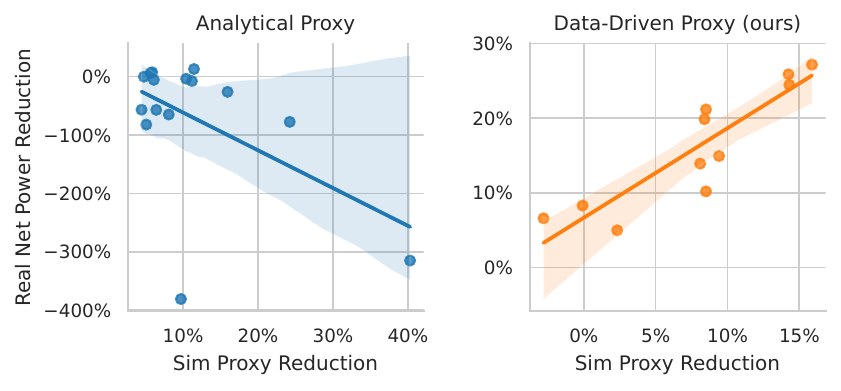}
\caption{Comparison of the effectiveness of the analytical proxy versus our data-driven model in net power reduction. The shaded area represents the confidence interval of the regression line.}
\vspace{-2.0em}
\label{fig:proxy-comparison}
\end{figure}

\paragraph{Data-driven vs. Analytical Proxy}
The role of the optimization proxy is critical in fine-tuning for factors that are challenging to simulate accurately. It serves as both the target for policy optimization and the criterion for initial policy selection. We compare the effectiveness of our data-driven model against the analytical proxy, focusing on their ability to accurately predict real-world power consumption.

Figure~\ref{fig:proxy-comparison} plots the power reductions measured in the real world against those predicted by both proxies. For the analytical proxy, we include all policies tested in the real world, noting that several were excluded due to deployment failure. The data-driven proxy data comprises the elite set $\Pi^\text{elites}$ from the last two iterations, using the corresponding measurement models as the proxy.

The data-driven proxy overall exhibits a positive correlation with actual power reductions observed in real-world operations, indicating its reliability and informativeness. In contrast, the analytical proxy generally shows a negative correlation, where many policies predicted to improve efficiency actually increase power consumption. This discrepancy highlights the shortcomings of the analytical proxy in capturing real-world dynamics, potentially leading to misguided policy optimization and selection.

Despite the general efficacy of the data-driven model, discrepancies between its predictions and the actual power reductions remain, highlighting the necessity for real-world testing of multiple policies to ascertain and retain the most effective policy. This requirement supports the incorporation of a hierarchical policy selection mechanism within our framework.

\section{Conclusions}
In this work, we have developed a fine-tuning methodology aimed at enhancing the performance of quadruped locomotion. This methodology specifically optimizes hard-to-simulate objectives by leveraging a data-driven measurement model. Our approach demonstrates a substantial reduction in total power consumption compared to traditional methods that rely on analytical proxies. Importantly, our framework requires minimal adaptations for integrating real-world knowledge into existing locomotion pipelines, enhancing its practical applicability.

Our study has several limitations. Algorithmically, the sim-to-real gap introduces out-of-distribution (OOD) challenges that compromise the reliability of our measurement model. Additionally, the framework requires a relatively large dataset of real-world samples to achieve convergence. Experimentally, due to hardware constraints, we were unable to test the transferability of our framework to other robot models. Furthermore, our experimental evaluation was limited to flat terrain, as measuring speed on more complex surfaces—such as slopes and stairs—presents additional challenges. Lastly, our focus was primarily on reducing total power consumption. Nevertheless, we are optimistic that, with minimal modifications, our framework could be extended to address other hard-to-simulate objectives.






\bibliographystyle{IEEEtran}
\bibliography{IEEEabrv, IEEEfull}{}

\begin{thebibliography}{10}
\providecommand{\url}[1]{#1}
\csname url@rmstyle\endcsname
\providecommand{\newblock}{\relax}
\providecommand{\bibinfo}[2]{#2}
\providecommand\BIBentrySTDinterwordspacing{\spaceskip=0pt\relax}
\providecommand\BIBentryALTinterwordstretchfactor{4}
\providecommand\BIBentryALTinterwordspacing{\spaceskip=\fontdimen2\font plus
\BIBentryALTinterwordstretchfactor\fontdimen3\font minus \fontdimen4\font\relax}
\providecommand\BIBforeignlanguage[2]{{%
\expandafter\ifx\csname l@#1\endcsname\relax
\typeout{** WARNING: IEEEtran.bst: No hyphenation pattern has been}%
\typeout{** loaded for the language `#1'. Using the pattern for}%
\typeout{** the default language instead.}%
\else
\language=\csname l@#1\endcsname
\fi
#2}}

\bibitem{ha2024learning}
S.~Ha, J.~Lee, M.~van~de Panne, Z.~Xie, W.~Yu, and M.~Khadiv, ``Learning-based legged locomotion; state of the art and future perspectives,'' \emph{arXiv preprint arXiv:2406.01152}, 2024.

\bibitem{biswal2021development}
P.~Biswal and P.~K. Mohanty, ``Development of quadruped walking robots: A review,'' \emph{Ain Shams Engineering Journal}, vol.~12, no.~2, pp. 2017--2031, 2021.

\bibitem{DUFFY2003177}
\BIBentryALTinterwordspacing
B.~R. Duffy, ``Anthropomorphism and the social robot,'' \emph{Robotics and Autonomous Systems}, vol.~42, no.~3, pp. 177--190, 2003, socially Interactive Robots. [Online]. Available: \url{https://www.sciencedirect.com/science/article/pii/S0921889002003743}
\BIBentrySTDinterwordspacing

\bibitem{breazeal2004designing}
C.~Breazeal, \emph{Designing sociable robots}.\hskip 1em plus 0.5em minus 0.4em\relax MIT press, 2004.

\bibitem{hagele2016industrial}
M.~H{\"a}gele, K.~Nilsson, J.~N. Pires, and R.~Bischoff, ``Industrial robotics,'' \emph{Springer handbook of robotics}, pp. 1385--1422, 2016.

\bibitem{unitreego1}
``Unitree go1,'' \url{https://www.unitree.com/go1}.

\bibitem{unitreego2}
``Unitree go2,'' \url{https://www.unitree.com/go2}.

\bibitem{katz2019mini}
B.~Katz, J.~Di~Carlo, and S.~Kim, ``Mini cheetah: A platform for pushing the limits of dynamic quadruped control,'' in \emph{2019 international conference on robotics and automation (ICRA)}.\hskip 1em plus 0.5em minus 0.4em\relax IEEE, 2019, pp. 6295--6301.

\bibitem{Trovato2018Sound}
G.~Trovato, R.~Paredes, J.~Balvin, F.~Cuellar, N.~B. Thomsen, S.~Bech, and Z.~Tan, ``The sound or silence: Investigating the influence of robot noise on proxemics,'' in \emph{2018 27th IEEE International Symposium on Robot and Human Interactive Communication (RO-MAN)}, 2018, pp. 713--718.

\bibitem{trujillo2009thermally}
S.~Trujillo and M.~Cutkosky, ``Thermally constrained motor operation for a climbing robot,'' in \emph{2009 IEEE International Conference on Robotics and Automation}.\hskip 1em plus 0.5em minus 0.4em\relax IEEE, 2009, pp. 2362--2367.

\bibitem{tan2018sim}
J.~Tan, T.~Zhang, E.~Coumans, A.~Iscen, Y.~Bai, D.~Hafner, S.~Bohez, and V.~Vanhoucke, ``Sim-to-real: Learning agile locomotion for quadruped robots,'' \emph{arXiv preprint arXiv:1804.10332}, 2018.

\bibitem{peng2018sim}
X.~B. Peng, M.~Andrychowicz, W.~Zaremba, and P.~Abbeel, ``Sim-to-real transfer of robotic control with dynamics randomization,'' in \emph{2018 IEEE international conference on robotics and automation (ICRA)}.\hskip 1em plus 0.5em minus 0.4em\relax IEEE, 2018, pp. 3803--3810.

\bibitem{hwangbo2019learning}
J.~Hwangbo, J.~Lee, A.~Dosovitskiy, D.~Bellicoso, V.~Tsounis, V.~Koltun, and M.~Hutter, ``Learning agile and dynamic motor skills for legged robots,'' \emph{Science Robotics}, vol.~4, no.~26, p. eaau5872, 2019.

\bibitem{xiong2024adaptive}
H.~Xiong, R.~Mendonca, K.~Shaw, and D.~Pathak, ``Adaptive mobile manipulation for articulated objects in the open world,'' \emph{arXiv preprint arXiv:2401.14403}, 2024.

\bibitem{agarwal2023legged}
A.~Agarwal, A.~Kumar, J.~Malik, and D.~Pathak, ``Legged locomotion in challenging terrains using egocentric vision,'' in \emph{Conference on robot learning}.\hskip 1em plus 0.5em minus 0.4em\relax PMLR, 2023, pp. 403--415.

\bibitem{margolis2023walk}
G.~B. Margolis and P.~Agrawal, ``Walk these ways: Tuning robot control for generalization with multiplicity of behavior,'' in \emph{Conference on Robot Learning}.\hskip 1em plus 0.5em minus 0.4em\relax PMLR, 2023, pp. 22--31.

\bibitem{zhuang2023robot}
Z.~Zhuang, Z.~Fu, J.~Wang, C.~Atkeson, S.~Schwertfeger, C.~Finn, and H.~Zhao, ``Robot parkour learning,'' \emph{arXiv preprint arXiv:2309.05665}, 2023.

\bibitem{margolis2024rapid}
G.~B. Margolis, G.~Yang, K.~Paigwar, T.~Chen, and P.~Agrawal, ``Rapid locomotion via reinforcement learning,'' \emph{The International Journal of Robotics Research}, vol.~43, no.~4, pp. 572--587, 2024.

\bibitem{peng2020learning}
X.~B. Peng, E.~Coumans, T.~Zhang, T.-W. Lee, J.~Tan, and S.~Levine, ``Learning agile robotic locomotion skills by imitating animals,'' \emph{arXiv preprint arXiv:2004.00784}, 2020.

\bibitem{escontrela2022adversarial}
A.~Escontrela, X.~B. Peng, W.~Yu, T.~Zhang, A.~Iscen, K.~Goldberg, and P.~Abbeel, ``Adversarial motion priors make good substitutes for complex reward functions,'' in \emph{2022 IEEE/RSJ International Conference on Intelligent Robots and Systems (IROS)}.\hskip 1em plus 0.5em minus 0.4em\relax IEEE, 2022, pp. 25--32.

\bibitem{isaacgym}
V.~Makoviychuk, L.~Wawrzyniak, Y.~Guo, M.~Lu, K.~Storey, M.~Macklin, D.~Hoeller, N.~Rudin, A.~Allshire, A.~Handa, and G.~State, ``Isaac gym: High performance gpu-based physics simulation for robot learning,'' 2021.

\bibitem{todorov2012mujoco}
E.~Todorov, T.~Erez, and Y.~Tassa, ``Mujoco: A physics engine for model-based control,'' \emph{2012 IEEE/RSJ International Conference on Intelligent Robots and Systems}, pp. 5026--5033, 2012.

\bibitem{coumans2019bullet}
E.~Coumans, Y.~Bai, D.~Hafner, V.~Vanhoucke, S.~Bohez, and J.~Tan, ``Bullet physics simulation: Recent developments and future directions,'' \emph{arXiv preprint arXiv:1904.00756}, 2019.

\bibitem{kundur2007power}
P.~Kundur, ``Power system stability,'' \emph{Power system stability and control}, vol.~10, pp. 7--1, 2007.

\bibitem{mellor2009computationally}
P.~Mellor, R.~Wrobel, and D.~Holliday, ``A computationally efficient iron loss model for brushless ac machines that caters for rated flux and field weakened operation,'' in \emph{2009 IEEE International Electric Machines and Drives Conference}.\hskip 1em plus 0.5em minus 0.4em\relax IEEE, 2009, pp. 490--494.

\bibitem{unitreego1motor}
``Unitree go1 motor,'' \url{https://www.unitree.com/go1motor/}.

\bibitem{wang2018advanced}
F.~Wang, Z.~Zhang, X.~Mei, J.~Rodr{\'\i}guez, and R.~Kennel, ``Advanced control strategies of induction machine: Field oriented control, direct torque control and model predictive control,'' \emph{energies}, vol.~11, no.~1, p. 120, 2018.

\bibitem{mahankali2024maximizing}
S.~Mahankali, C.-C. Lee, G.~B. Margolis, Z.-W. Hong, and P.~Agrawal, ``Maximizing quadruped velocity by minimizing energy,'' \emph{International Conference on Robotics and Automation}, 2024.

\bibitem{yang2022fast}
Y.~Yang, T.~Zhang, E.~Coumans, J.~Tan, and B.~Boots, ``Fast and efficient locomotion via learned gait transitions,'' in \emph{Conference on robot learning}.\hskip 1em plus 0.5em minus 0.4em\relax PMLR, 2022, pp. 773--783.

\bibitem{fu2021minimizing}
Z.~Fu, A.~Kumar, J.~Malik, and D.~Pathak, ``Minimizing energy consumption leads to the emergence of gaits in legged robots,'' \emph{arXiv preprint arXiv:2111.01674}, 2021.

\bibitem{rudin2022learning}
N.~Rudin, D.~Hoeller, P.~Reist, and M.~Hutter, ``Learning to walk in minutes using massively parallel deep reinforcement learning,'' in \emph{Conference on Robot Learning}.\hskip 1em plus 0.5em minus 0.4em\relax PMLR, 2022, pp. 91--100.

\bibitem{minicheetahbattery}
``Robotsguide: Mini cheetah,'' \url{https://robotsguide.com/robots/minicheetah}.

\bibitem{seok2014design}
S.~Seok, A.~Wang, M.~Y. Chuah, D.~J. Hyun, J.~Lee, D.~M. Otten, J.~H. Lang, and S.~Kim, ``Design principles for energy-efficient legged locomotion and implementation on the mit cheetah robot,'' \emph{Ieee/asme transactions on mechatronics}, vol.~20, no.~3, pp. 1117--1129, 2014.

\bibitem{ha2020learning}
S.~Ha, P.~Xu, Z.~Tan, S.~Levine, and J.~Tan, ``Learning to walk in the real world with minimal human effort,'' \emph{arXiv preprint arXiv:2002.08550}, 2020.

\bibitem{haarnoja2018learning}
T.~Haarnoja, S.~Ha, A.~Zhou, J.~Tan, G.~Tucker, and S.~Levine, ``Learning to walk via deep reinforcement learning,'' \emph{arXiv preprint arXiv:1812.11103}, 2018.

\bibitem{smith2022walk}
L.~Smith, I.~Kostrikov, and S.~Levine, ``A walk in the park: Learning to walk in 20 minutes with model-free reinforcement learning,'' \emph{arXiv preprint arXiv:2208.07860}, 2022.

\bibitem{smith2023grow}
L.~Smith, Y.~Cao, and S.~Levine, ``Grow your limits: Continuous improvement with real-world rl for robotic locomotion,'' \emph{arXiv preprint arXiv:2310.17634}, 2023.

\bibitem{smith2022legged}
L.~Smith, J.~C. Kew, X.~B. Peng, S.~Ha, J.~Tan, and S.~Levine, ``Legged robots that keep on learning: Fine-tuning locomotion policies in the real world,'' in \emph{2022 International Conference on Robotics and Automation (ICRA)}.\hskip 1em plus 0.5em minus 0.4em\relax IEEE, 2022, pp. 1593--1599.

\bibitem{lei2023uni}
K.~Lei, Z.~He, C.~Lu, K.~Hu, Y.~Gao, and H.~Xu, ``Uni-o4: Unifying online and offline deep reinforcement learning with multi-step on-policy optimization,'' \emph{arXiv preprint arXiv:2311.03351}, 2023.

\bibitem{shi2024efficient}
H.~Shi, T.~Li, Q.~Zhu, J.~Sheng, L.~Han, and M.~Q.-H. Meng, ``An efficient model-based approach on learning agile motor skills without reinforcement,'' \emph{arXiv preprint arXiv:2403.01962}, 2024.

\bibitem{hiraoka2021dropout}
T.~Hiraoka, T.~Imagawa, T.~Hashimoto, T.~Onishi, and Y.~Tsuruoka, ``Dropout q-functions for doubly efficient reinforcement learning,'' \emph{arXiv preprint arXiv:2110.02034}, 2021.

\bibitem{shao2020controlvae}
H.~Shao, S.~Yao, D.~Sun, A.~Zhang, S.~Liu, D.~Liu, J.~Wang, and T.~Abdelzaher, ``Controlvae: Controllable variational autoencoder,'' in \emph{International conference on machine learning}.\hskip 1em plus 0.5em minus 0.4em\relax PMLR, 2020, pp. 8655--8664.

\bibitem{heiden2021neuralsim}
E.~Heiden, D.~Millard, E.~Coumans, Y.~Sheng, and G.~S. Sukhatme, ``Neuralsim: Augmenting differentiable simulators with neural networks,'' in \emph{2021 IEEE International Conference on Robotics and Automation (ICRA)}.\hskip 1em plus 0.5em minus 0.4em\relax IEEE, 2021, pp. 9474--9481.

\bibitem{qiao2021efficient}
Y.-L. Qiao, J.~Liang, V.~Koltun, and M.~C. Lin, ``Efficient differentiable simulation of articulated bodies,'' in \emph{International Conference on Machine Learning}.\hskip 1em plus 0.5em minus 0.4em\relax PMLR, 2021, pp. 8661--8671.

\bibitem{heiden2022inferring}
E.~Heiden, Z.~Liu, V.~Vineet, E.~Coumans, and G.~S. Sukhatme, ``Inferring articulated rigid body dynamics from rgbd video,'' in \emph{2022 IEEE/RSJ International Conference on Intelligent Robots and Systems (IROS)}.\hskip 1em plus 0.5em minus 0.4em\relax IEEE, 2022, pp. 8383--8390.

\bibitem{sanchez2020learning}
A.~Sanchez-Gonzalez, J.~Godwin, T.~Pfaff, R.~Ying, J.~Leskovec, and P.~Battaglia, ``Learning to simulate complex physics with graph networks,'' in \emph{International conference on machine learning}.\hskip 1em plus 0.5em minus 0.4em\relax PMLR, 2020, pp. 8459--8468.

\bibitem{ajay2018augmenting}
A.~Ajay, J.~Wu, N.~Fazeli, M.~Bauza, L.~P. Kaelbling, J.~B. Tenenbaum, and A.~Rodriguez, ``Augmenting physical simulators with stochastic neural networks: Case study of planar pushing and bouncing,'' in \emph{2018 IEEE/RSJ International Conference on Intelligent Robots and Systems (IROS)}.\hskip 1em plus 0.5em minus 0.4em\relax IEEE, 2018, pp. 3066--3073.

\bibitem{kaufmann2023champion}
E.~Kaufmann, L.~Bauersfeld, A.~Loquercio, M.~M{\"u}ller, V.~Koltun, and D.~Scaramuzza, ``Champion-level drone racing using deep reinforcement learning,'' \emph{Nature}, vol. 620, no. 7976, pp. 982--987, 2023.

\bibitem{abeyruwan2023sim2real}
S.~W. Abeyruwan, L.~Graesser, D.~B. D’Ambrosio, A.~Singh, A.~Shankar, A.~Bewley, D.~Jain, K.~M. Choromanski, and P.~R. Sanketi, ``i-sim2real: Reinforcement learning of robotic policies in tight human-robot interaction loops,'' in \emph{Conference on Robot Learning}.\hskip 1em plus 0.5em minus 0.4em\relax PMLR, 2023, pp. 212--224.

\bibitem{krimsky2024elastic}
E.~Krimsky and S.~H. Collins, ``Elastic energy-recycling actuators for efficient robots,'' \emph{Science Robotics}, vol.~9, no.~88, p. eadj7246, 2024.

\bibitem{kim2024not}
Y.~Kim, H.~Oh, J.~Lee, J.~Choi, G.~Ji, M.~Jung, D.~Youm, and J.~Hwangbo, ``Not only rewards but also constraints: Applications on legged robot locomotion,'' \emph{IEEE Transactions on Robotics}, 2024.

\bibitem{chane2024cat}
E.~Chane-Sane, P.-A. Leziart, T.~Flayols, O.~Stasse, P.~Sou{\`e}res, and N.~Mansard, ``Cat: Constraints as terminations for legged locomotion reinforcement learning,'' \emph{arXiv preprint arXiv:2403.18765}, 2024.

\bibitem{sener2018multi}
O.~Sener and V.~Koltun, ``Multi-task learning as multi-objective optimization,'' \emph{Advances in neural information processing systems}, vol.~31, 2018.

\bibitem{gunantara2018review}
N.~Gunantara, ``A review of multi-objective optimization: Methods and its applications,'' \emph{Cogent Engineering}, vol.~5, no.~1, p. 1502242, 2018.

\bibitem{schmitt2018kickstarting}
S.~Schmitt, J.~J. Hudson, A.~Zidek, S.~Osindero, C.~Doersch, W.~M. Czarnecki, J.~Z. Leibo, H.~Kuttler, A.~Zisserman, K.~Simonyan, \emph{et~al.}, ``Kickstarting deep reinforcement learning,'' \emph{arXiv preprint arXiv:1803.03835}, 2018.

\bibitem{nikishin2022primacy}
E.~Nikishin, M.~Schwarzer, P.~D’Oro, P.-L. Bacon, and A.~Courville, ``The primacy bias in deep reinforcement learning,'' in \emph{International conference on machine learning}.\hskip 1em plus 0.5em minus 0.4em\relax PMLR, 2022, pp. 16\,828--16\,847.

\bibitem{schwarzer2023bigger}
M.~Schwarzer, J.~S.~O. Ceron, A.~Courville, M.~G. Bellemare, R.~Agarwal, and P.~S. Castro, ``Bigger, better, faster: Human-level atari with human-level efficiency,'' in \emph{International Conference on Machine Learning}.\hskip 1em plus 0.5em minus 0.4em\relax PMLR, 2023, pp. 30\,365--30\,380.

\bibitem{naik2024reward}
A.~Naik, Y.~Wan, M.~Tomar, and R.~S. Sutton, ``Reward centering,'' \emph{arXiv preprint arXiv:2405.09999}, 2024.

\bibitem{schulman2017proximal}
J.~Schulman, F.~Wolski, P.~Dhariwal, A.~Radford, and O.~Klimov, ``Proximal policy optimization algorithms,'' \emph{arXiv preprint arXiv:1707.06347}, 2017.

\end{thebibliography}

\end{document}